\documentclass[letterpaper, 10 pt, conference]{ieeeconf}

\IEEEoverridecommandlockouts                              

\usepackage{graphicx}
\graphicspath{{figs/}}
\usepackage{amsmath}
\usepackage{amssymb}
\usepackage{algorithm}
\usepackage{algpseudocode}
\usepackage{times}
\usepackage{bm}
\usepackage{color}
\usepackage{mathtools}
\usepackage{algorithm}
\usepackage{algpseudocode}
\usepackage{multirow}
\usepackage{hyperref}
\newcommand{\figref}[1]{Fig.~\ref{#1}}

\newcommand{\secref}[1]{Section~\ref{#1}}

\newcommand{\R}{\mathbb{R}}

\newcommand{\SO}{\mathop{\mathrm{SO}}}

\newcommand{\qt}{\bm{q}}

\newcommand{\Sp}{\mathbb{S}}
\newcommand{\bingham}{\mathfrak{B}}

\newcommand{\Expect}[1]{\operatorname{\mathcal{E}}\{#1\}}
\newcommand{\trace}{\operatorname{tr}}

\newcommand{\vect}[3]{{}^{#2}\hspace{-.3ex}{#1}_{#3}}
\newcommand{\widehatvect}[3]{{}^{#2}\hspace{-.3ex}\widehat{#1}_{#3}}
\newcommand{\dvect}[3]{{}^{#2}\hspace{-.3ex}{\dot{#1}}_{#3}}
\newcommand{\ddvect}[3]{{}^{#2}\hspace{-.3ex}{\ddot{#1}}_{#3}}

\newcommand{\rvec}{\boldsymbol{r}}
\newcommand{\gyro}{\boldsymbol{\omega}}
\newcommand{\cm}[1]{\left[#1 \times\right]}

\newcommand{\dt}[1]{\frac{\mathrm{d}#1}{\mathrm{d}t}}

\newcommand{\frm}[1]{\{#1\}}

\title{\LARGE \bf
A Robot Kinematics Model Estimation Using Inertial Sensors \\ for On-Site Building Robotics
}

\author{Hiroya Sato$^\text{1}$, Tasuku Makabe$^\text{1}$, Iori Yanokura$^\text{1}$, Naoya Yamaguchi$^\text{1}$, Kei Okada$^\text{1}$ and Masayuki Inaba$^\text{1}$
\thanks{$^{\text{1}}$Authors are with Department of Mechano-Informatics, Graduate School
of Information Science and Technology, The University of Tokyo, 7-3-1
Hongo, Bunkyo-ku, Tokyo, 113-8656, Japan.
{\tt\footnotesize [h-sato, makabe, yanokura, yamaguchi, k-okada, inaba]@jsk.t.u-tokyo.ac.jp}
}%
}%

\begin{document}

\maketitle
\thispagestyle{empty}
\pagestyle{empty}

\begin{abstract}


In order to make robots more useful in a variety of environments, they need to be highly portable so that they can be transported to wherever they are needed, and highly storable so that they can be stored when not in use. We propose ``on-site robotics'', which uses parts procured at the location where the robot will be active,
and propose a new solution to the problem of portability and storability.
In this paper, as a proof of concept for on-site robotics, we describe a method for estimating the kinematic model of a robot by using inertial measurement units (IMU) sensor module on rigid links, estimating the relative orientation between modules from angular velocity, and estimating the relative position from the measurement of centrifugal force.

At the end of this paper, as an evaluation for this method, we present an experiment in which a robot made up of wooden sticks reaches a target position. In this experiment, even if the combination of the links is changed, the robot is able to reach the target position again immediately after estimation, showing that it can operate even after being reassembled.
%
Our implementation is available on \url{https://github.com/hiroya1224/urdf_estimation_with_imus}.
\end{abstract}

\section{INTRODUCTION}

In today's robotics, it is usually assumed to use well-designed robots whose dynamics and kinematics have been analyzed as accurately as possible.
Therefore, when a robot is transported, the analyzed material must also be transported.
Even if the robot can be disassembled or deformed, the total mass remains constant, so the larger the robot, the higher the transportation cost. In addition to transportation issues, how to store the robots is another practical problem.

One possible solution to this problem is soft robotics. The use of lightweight or shape-deformable materials can improve portability and storability.
However, as is currently being actively discussed, controlling highly flexible links is difficult, and the development of new control methods to replace conventional rigid robots is a challenge.

\begin{figure}[t]
    \centering
    \includegraphics[width=\linewidth]{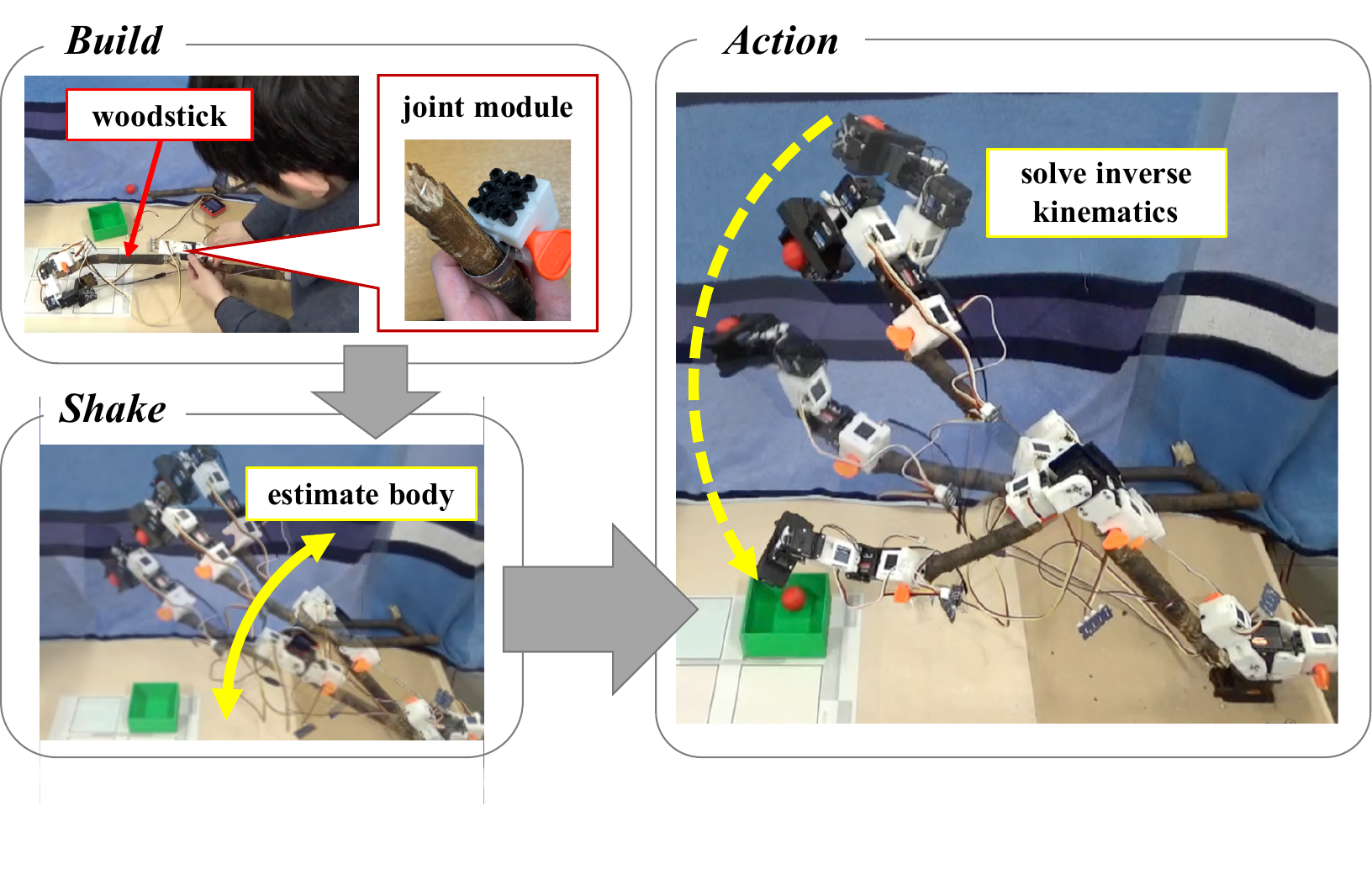}
    \vspace{-1.25cm}
    \caption{Sequence of kinematic model estimation. In an experiment to evaluate our method, we used centrifugal force to estimate the length of a link by shaking a manipulator made of wooden sticks. By using the link length estimation results, the inverse kinematics of the gripper can be solved toward the target position.}
    \label{fig:concept}
    \vspace{-0.33cm}
\end{figure}

As an alternative approach to solving the portability and storage problem, we propose a ``on-site building robotics'', in which the robot is assembled with locally procured materials. This robotics improves the portability of the robot while effectively utilizing conventional rigid-body robot methods. Only the joint intermediate modules need to be kept on hand, and link parts that are no longer needed can be discarded, thereby improving storability.

However, there are many challenges to realizing such a robotics.
In particular, if the robot is allowed to be composed of arbitrary rigid components, it must solve a hard problem of estimating its own robot model, which is necessary to realize the motion. 
Depending on the type of rigid rod used, it may not be straight but curved complexly. In such cases, using a ruler, for example, it is not easy to measure the relative pose between the joints, especially the relative orientation.
When trying to estimate relative pose using an external camera and markers, it is still difficult because the longer the link, the harder it is to keep everything within the same field of view.

To generate a complete robot model, both kinematic and dynamic parameters are needed. According to the classical argument \cite{inertiaestim_1986}, dynamic parameters such as mass and moment of inertia can be linearly separated using matrices containing link lengths and joint angles in nonlinear form. This implies that if the relative poses of the joints and the relative angles of the joints are known, the dynamics parameters can be estimated relatively easily.

This paper focuses on the estimation of robot kinematic model using joint intermediate modules as a proof of concept for on-site building robotics.
We propose a novel method for estimating the relative pose of joints using multiple inertial measurement unit (IMU) sensor modules.
At the end of the paper, an evaluation experiment on a manipulator made of wooden sticks is presented to validate the effectiveness of the method.




\section{RELATED WORKS}

\subsection{Soft Robotics as a Solution for Portability}

Inflatable robots are being developed for their portability and ability to work in confined spaces.
A classical example is the inflatable manipulator proposed by \cite{RYBSKI1996111}.
Inflatable mobility is proposed in \cite{inflatable_mobility}, which is a new vehicle with high portability and low impact mitigation in the event of a collision.
An alternative approach is to use origami robotics \cite{origami_manipulator, foldable2017}, which can be folded for compactness.

\subsection{Link Length Estimation in Robot Calibration}

Estimation of kinematic parameters of robots has been widely studied in the field of calibration \cite{robot_calib, mooring1991fundamentals}.
When the body parameters are known to some extent, methods have been proposed to correct the body parameters with small corrections, assuming that the error from the true value is small. Such methods have been used to improve the dexterity of manipulators.

In \cite{Ye_2006, BALANJI2022102248}, a method for correcting body parameters based on the pose of the target frame was proposed by defining a differential model. 
For multiple joint angles, the target frame calculated from the kinematic model and the actual frame pose observed by an external optical sensor are collected. The error term is then corrected by the method of least squares.

These methods assume the existence of an external optical sensor to take the true value and that the kinematic model of the robot is approximately known in advance.

\subsection{Acquisition of Self-Body Schema and Tool Usages}

In that the robots manipulate objects whose length is not known in advance, the context of tool use is similar.
The use of tools by robots is often discussed in conjunction with the acquisition of a robot's self-body schema.
The concept of body schema is a term in cognitive neuroscience, which originally refers to a human's self-understanding of its own body, and is often cited in robotics \cite{bodyschema_hoffmann}.

In \cite{Sturm2012}, they proposed a learning process for a kinematic model of a robot whose model is completely unknown based on the position of AR markers attached to the links.
In this study, the joint relationship between joints and links was estimated by providing a large amount of data of link pose.
In \cite{adaptivebodyschema_tool}, a neural network method was proposed to learn a tool of unknown length using information from tactile sensors.

This is similar to our work in that it uses only internal sensors. However, our proposal allows additional modules to be placed at the end of the link, and it is possible to collect data that is advantageous for estimation. We approach the problem in a simpler way that does not require prior learning.

\subsection{Robot Prototyping and Modularized Robot}

Robot prototyping research is usually concerned with the creation of robot components \cite{rapid_proto,fdm_proto2022}.
%
In the context of educational robotics, there is an example of building a robot with an actuated paper craft capable of inductive power transfer \cite{autogami}.
They claimed to be useful for robot prototyping thanks to its ease of assembly.
Although it is actuated by a shape memory alloy (SMA), its advanced behaviors, such as inverse kinematics, are restricted.

A modular robot is a method of constructing a robot body by repeatedly using the same parts \cite{modularrobot}.
Since the same elements are used repeatedly, a kinematic model can be obtained in principle if the connection structure is known.
However, there is a strong restriction that only one module can be used, and adding a new module requires sacrificing the advantage of repetition of the same elements.

One commercially available robot prototypable product is \cite{studica_prototyping}, which allows for easy robot assembly.
It provides a set of easy to assemble robot bodies by combining well-designed robot modules.
In contrast, our goal is to immediately assemble and operate the robot using materials available on site, which is different from the case where the components to be used are known in advance.

\subsection{Relative Attitude Estimation of IMUs}

The method of estimating relative pose of IMUs are well conducted in the field of clinical application.
There have been proposed the method \cite{kok_imu, Weygers_driftfree} for gait analysis under the assumption that the relative position of joint (in their case, knees) with respect to IMU sensors are known.
Our approach is very similar to this study, but it is not directly applicable, because their method has limitations in its use for relative pose estimation of IMUs whose relative positions are completely unknown.

\section{OUR METHOD}
\begin{figure}[t]
    \centering
    \includegraphics[width=\linewidth]{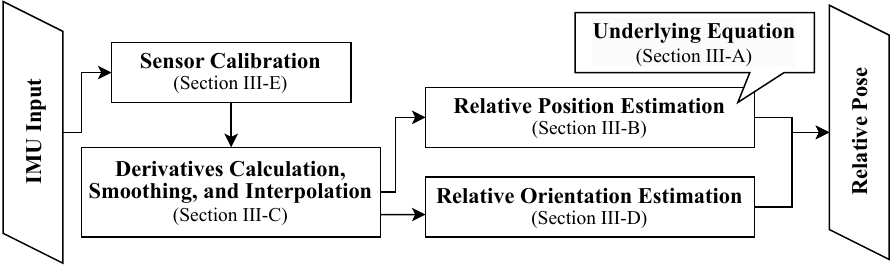}
    \caption{Flowchart of our proposed relative pose estimation method. Corresponding sections are shown in each box.}
    \vspace{-0.25cm}
    \label{fig:flowchart}
\end{figure}

We propose the method of estimating joints' relative pose based on the acceleration and angular velocity observed by two IMUs.
An overview of our method is shown in \figref{fig:flowchart}.

\subsection{Equations of Motion for IMU Sensors}

If $\vect{\bm{r}}{O}{A}$ and $\vect{\bm{r}}{O}{P}$ denotes the position of the spatial points $A$, $P$ in the 3D Euclidean space with respect to the coordinate frame $O$, respectively, then the following equation holds: 
\begin{equation}
    \vect{\rvec}{O}{P} = \vect{\rvec}{O}{A} + \vect{R}{O}{A} \vect{\rvec}{A}{P}, \label{eq:relpos}
\end{equation}
where $\vect{R}{O}{A}$ denotes the relative rotation of the coordinate frame $\frm{A}$ with respect to $\frm{O}$, and $\vect{\rvec}{A}{P}$ the relative position of $P$ with respect to $A$. By differentiate twice \eqref{eq:relpos} with time, we get
\begin{align}
    \ddvect{\rvec}{O}{P}
    &= \ddvect{\rvec}{O}{A} + \vect{R}{O}{A} \left( \cm{\vect{\gyro}{A}{A}}^2 + \cm{\dvect{\gyro}{A}{A}} \right) \vect{\rvec}{A}{P} \nonumber \\
    &+ 2 \left.\vect{R}{O}{A} \cm{\vect{\gyro}{A}{A}} \dvect{\rvec}{A}{P}\right. + \vect{R}{O}{A} \ddvect{\rvec}{A}{P} \label{eq:relacc}
\end{align}
Here we used the time derivative of rotation matrix:
\begin{equation}
    \dt{\vect{R}{O}{A}} = \vect{R}{O}{A} \cm{\vect{\gyro}{A}{A}}.
\end{equation}

Now we assume that the relative motion between $A$ and $P$ is negelible, i.e., $\dvect{\rvec}{A}{P} = \ddvect{\rvec}{A}{P} = \bm{0}$. Then \eqref{eq:relacc} reduced to
\begin{equation}
    \ddvect{\rvec}{O}{P}
    = \ddvect{\rvec}{O}{A} + \vect{R}{O}{A} \left( \cm{\vect{\gyro}{A}{A}}^2 + \cm{\dvect{\gyro}{A}{A}} \right) \vect{\rvec}{A}{P}. \label{eq:reduced_relacc}
\end{equation}

Letting $\vect{\bm{f}}{O}{A}$ be a specific force, i.e., the force divide by the object's mass, acting on the origin of the frame $\frm{A}$ representing by the basis of the frame $\frm{O}$, then \eqref{eq:reduced_relacc} becomes as follows by the Newton's law.
\begin{equation}
    \vect{\bm{f}}{O}{P} - \vect{\bm{f}}{O}{A} = \vect{R}{O}{A} \left( \cm{\vect{\gyro}{A}{A}}^2 + \cm{\dvect{\gyro}{A}{A}} \right) \vect{\rvec}{A}{P}. \label{eq:forceform}
\end{equation}

IMUs yield the specific force vector representating by the basis of their base origin. In other words, the acceleration data yielding by the IMUs are in the form $\vect{\bm{f}}{A}{A}$ or $\vect{\bm{f}}{P}{P}$. 
To obtain them, multiply from the left by $\vect{R}{A}{O} = \vect{R}{O}{A}^\top$. 
\begin{equation}
    \vect{R}{A}{P}\vect{\bm{f}}{P}{P} - \vect{\bm{f}}{A}{A}
    = \left( \cm{\vect{\gyro}{A}{A}}^2 + \cm{\dvect{\gyro}{A}{A}} \right) \vect{\rvec}{A}{P}. \label{eq:relposition_estimator}
\end{equation}
If we already know $\vect{R}{A}{P}$, the relative rotation of frame $\frm{P}$ with respect to $\frm{A}$, then \eqref{eq:relposition_estimator} becomes the simple linear equation with unknown variable $\vect{\rvec}{A}{P}$, which is what we want.
We define
\begin{equation}
    \Omega(\gyro) = \cm{\gyro}^2 + \cm{\dot{\gyro}}.
\end{equation}
After an easy algebra, we get
\begin{align}
    \vect{R}{A}{P} \vect{\bm{f}}{P}{P} - \vect{\bm{f}}{A}{A}
    &= \vect{R}{A}{P} \Omega\left(\vect{\gyro}{P}{P}\right) \vect{R}{A}{P}^\top \vect{\rvec}{A}{P}.
\end{align}
Taking average with \eqref{eq:relposition_estimator}, we finally yield
\begin{align}
    \bm{F}(\vect{\bm{f}}{A}{A},\vect{\bm{f}}{P}{P})
    &= \overline{\Omega}(\vect{\gyro}{A}{A}, \vect{\gyro}{P}{P}) \vect{\rvec}{A}{P}, \label{eq:estimate_equation}
\end{align}
where
\begin{align}
    \bm{F}(\vect{\bm{f}}{A}{A},\vect{\bm{f}}{P}{P}) &= \vect{R}{A}{P}\vect{\bm{f}}{P}{P} - \vect{\bm{f}}{A}{A}, \label{eq:vecF_motion} \\
    \overline{\Omega}(\vect{\gyro}{A}{A}, \vect{\gyro}{P}{P}) &= \frac{1}{2}\left(\Omega\left(\vect{\gyro}{A}{A}\right) + \vect{R}{A}{P} \Omega\left(\vect{\gyro}{P}{P}\right)\vect{R}{A}{P}^\top \right). \label{eq:matX_motion}
\end{align}

\subsection{Recursive Least Square}
\begin{algorithm}[t]
    \caption{RLS for Relative Position Estimation}
    \label{alg:ourmethod}
    {
    \setlength{\baselineskip}{1.2\baselineskip}
    \begin{algorithmic}[1]











    \State $P_{\text{inv}} \gets \varepsilon I_3$ \Comment{$\varepsilon > 0$ is sufficiently small value and $I_n$ the identity matrix of size $n$}
    \State $\widehatvect{\rvec}{A}{P} \gets \boldsymbol{0}_3 = (0,0,0)^\top$
    \State Set $\Sigma^{\vect{\gyro}{A}{A}}, \Sigma^{\vect{\gyro}{P}{P}}$ \Comment{obtain during the calibration step described in \secref{sec:calibration}}
    \State $\mathtt{Diffs} = [\boldsymbol{0}_3, \dots, \boldsymbol{0}_3]$ \Comment{queue of length $N_\text{lag}$}
    
    \Function {RLS}{$
        \vect{\bm{f}}{A}{A}, 
        \vect{\bm{f}}{P}{P}, 
        \vect{\gyro}{A}{A}, 
        \vect{\gyro}{P}{P} ; \widehatvect{\bm{r}}{A}{P}, \vect{R}{A}{P} 
    $}
        \State Calculate $\bm{F}, \overline{\Omega}_0$ \Comment{Eqs. \eqref{eq:Fk}, \eqref{eq:Omega_zero}}

        \State Enqueue $\bm{F} - \overline{\Omega}_0 \widehatvect{\bm{r}}{A}{P}$ to $\mathtt{Diffs}$
        \State Dequeue the front entry of $\mathtt{Diffs}$

        \State $C \gets \left(\Sigma_{\bm{\delta} \in \mathtt{Diffs}} \, \bm{\delta} \bm{\delta}^\top
        \right) / (N_\text{lag} - 1)$ \Comment{Eq. \eqref{eq:Cn}}

        \State $P_\text{inv} \gets P_\text{inv} + \gamma_\text{pos} \cdot \overline{\Omega}_0^\top C^{-1} \overline{\Omega}_0$ \Comment{Eq. \eqref{eq:P_sequential}}

        \State $\bm{Q} \gets \bm{Q} + \gamma_\text{pos} \cdot \overline{\Omega}_0^{\top} C^{-1} \bm{F}$ \Comment{Eq. \eqref{eq:Q_sequential}}

        \State $\widehatvect{\rvec}{A}{P} \gets P_\text{inv}^{-1} \bm{Q}, \quad \widehat{\Sigma}^{\vect{\rvec}{A}{P}} \gets P_\text{inv}^{-1}$
        \State \Return $\widehatvect{\rvec}{A}{P}, \widehat{\Sigma}^{\vect{\rvec}{A}{P}}$
    \EndFunction

    \While{sensors are available} 
        \State $
        \vect{\bm{f}}{A}{A}, 
        \vect{\bm{f}}{P}{P}, 
        \vect{\gyro}{A}{A}, 
        \vect{\gyro}{P}{P} \gets \textsc{Calibrated\_IMU\_Data}$
        \State $
        \vect{R}{A}{P} \gets \text{mode of } \bingham(A)$ \Comment{Eq. \eqref{eq:binghammatrix} in \secref{sec:binghamfilter}}
        \State $\widehatvect{\rvec}{A}{P}, \widehat{\Sigma}^{\vect{\rvec}{A}{P}} \gets \textsc{RLS}(
        \vect{\bm{f}}{A}{A}, 
        \vect{\bm{f}}{P}{P}, 
        \vect{\gyro}{A}{A}, 
        \vect{\gyro}{P}{P}; \widehatvect{\bm{r}}{A}{P}, \vect{R}{A}{P} )$
    \EndWhile
    \end{algorithmic}
    }
\end{algorithm}

To achieve online estimation of $\vect{\rvec}{A}{P}$,
we adopt the matrix version of a recursive least square (RLS) method (see the Section 4.11.1 in \cite{moon2000mathematical}). 
We set the observed $n$ IMU data as
\begin{equation}
    \{(\vect{\bm{f}}{A}{A}^{(k)}, \vect{\bm{f}}{P}{P}^{(k)}, \vect{\gyro}{A}{A}^{(k)}, \vect{\gyro}{P}{P}^{(k)})\}_{k=1}^{n}.
\end{equation}
Let us consider the least square estimation of $\vect{\rvec}{A}{P}$ based on \eqref{eq:estimate_equation}.
Unfortunately, because $\Omega(\gyro)$ contains a second-order term of angular velocity, the mean of the residuals $\Expect{\bm{F} - \overline{\Omega} \vect{\rvec}{A}{P}}$ becomes non-zero, which violates the assumption of the least square regression.
To make the mean of the residuals zero, we add the correction term $K_{\overline{\Omega}}$ defined below to $\overline{\Omega}$.
\begin{equation}
    \begin{aligned}
    K_{\overline{\Omega}}(\Sigma^{\vect{\gyro}{A}{A}}, \Sigma^{\vect{\gyro}{P}{P}}) &= \frac{1}{2} \left[\trace\left(\Sigma^{\vect{\gyro}{A}{A}} + \Sigma^{\vect{\gyro}{P}{P}}\right) \cdot I_3 \right.\\
    &\left.- \left(\Sigma^{\vect{\gyro}{A}{A}}+ \vect{R}{A}{P} \,\Sigma^{\vect{\gyro}{P}{P}}\, \vect{R}{A}{P}^\top\right)\right].
    \end{aligned}
\end{equation}
For each sample, we calculate the vector $\bm{F}$ and $\overline{\Omega}_0 \, ( = \overline{\Omega} + K_{\overline{\Omega}})$ using \eqref{eq:vecF_motion} and \eqref{eq:matX_motion}.
\begin{align}
    \bm{F}^{(k)} &= \vect{R}{A}{P}\vect{\bm{f}}{P}{P}^{(k)} - \vect{\bm{f}}{A}{A}^{(k)}, \label{eq:Fk}\\
    \overline{\Omega}_0^{(k)} &= \overline{\Omega}(\vect{\gyro}{A}{A}^{(k)}, \vect{\gyro}{P}{P}^{(k)}) + K_{\overline{\Omega}}(\Sigma^{\vect{\gyro}{A}{A}}, \Sigma^{\vect{\gyro}{P}{P}}) \label{eq:Omega_zero}
\end{align}

Now we try to find the minimizer $\vect{\rvec}{A}{P}$ of $J[n]$ defined by the following:
\begin{equation}
J[n]= \frac{1}{2} \sum_{k=0}^n \gamma_\text{pos}^{n-k} \left\|\bm{F}^{(k)} - \overline{\Omega}_0^{(k)} \vect{\rvec}{A}{P} \right\|^2,
\end{equation}
%
where $\gamma_\text{pos} \in (0,1]$ is the forgetting factor.
The optimized
solution $\widehatvect{\rvec}{A}{P}[n]$ that minimizes $J[n]$ is written as follows.
\begin{equation}
    \widehatvect{\rvec}{A}{P}[n]=P[n]\bm{Q}[n], \label{eq:minimizer}
\end{equation}
where
\begin{align}
    P[n]^{-1} 
    & = P[n-1]^{-1} + \gamma_\text{pos} \cdot \overline{\Omega}_0^{(n)^{\top}} C[n]^{-1} \overline{\Omega}_0^{(n)}, \label{eq:P_sequential} \\
    \boldsymbol{Q}[n]
    & = \boldsymbol{Q}[n-1] + \gamma_\text{pos} \cdot \overline{\Omega}_0^{(n){\top}} C[n]^{-1} \bm{F}^{(n)}. \label{eq:Q_sequential}
\end{align}
Letting $\boldsymbol{\delta}^{(i)}[k] = \bm{F}^{(i)} - \overline{\Omega}_0^{(i)} \widehatvect{\rvec}{A}{P}[k]$, we define $C[n]$ as below.
\begin{equation}
    C[n] = \frac{1}{N_\text{lag} - 1} \sum_{i=n-N_\text{lag}}^n 
    \boldsymbol{\delta}^{(i)}[n-1] \boldsymbol{\delta}^{(i)}[n-1]^\top, \label{eq:Cn}
\end{equation}
where $\boldsymbol{\delta}^{(i)}$ is set to zero if $i < 0$.
$N_\text{lag}$, a sample time of estimation error, is an arbitrary positive integer larger than two. We set $N_\text{lag} = 100$. 
Note that if the number of non-zero $\boldsymbol{\delta}^{(i)}$ in the summation is less than 3, then $C[n]$ is always singular matrix. In this case, we use Moore-Penrose pseudoinverse instead of the ordinal inverse $C[n]^{-1}$. This step is needed only in the beginning of the estimation.
%
%


\begin{figure}[t]
    \centering
    \includegraphics[width=0.75\linewidth]{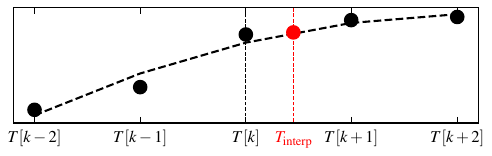}
    \caption{Time interpolation using Savitzky-Golay filter with $N=2$. If we want to align with the value at time $T_\text{interp}$, we can use polynomial interpolation to estimate the value at that time. If $T_\text{interp}$ falls outside the interpolation range, the closer of the values of $T[k \pm N]$ is used instead.}
    \vspace{-0.25cm}
    \label{fig:interpolation}
\end{figure}




\subsection{Savitzky-Golay Filter}

To estimate the relative position of IMUs from the centrifugal force, it is necessary to obtain a term for the time derivative of the angular velocity.
We calculated the time derivative using a Savitzky-Golay filter \cite{sgfilter_freq}.
It
is achieved by least-squares fitting of a polynomial of degree $M$ to the local $2N+1$ points, where $M$ and $N$ are hyperparameters to be determined. We use $M=5$ and $2N+1 = 7$ in the experiments. 
Details are discussed in \secref{sec:freqanalysis}.
%

We now consider applying a filter to the observed data sequence $\{(T[i], d[i])\}_i$.
The symbol $d$ is substituted for each of the acceleration and angular velocity terms.
The approximated polynomial can be written as follows.
\begin{equation}
    p[k](t) = \sum_{i=0}^{M} c_{k,i} (t - T[k])^i,
\end{equation}
where $c_{k,i} \,(i=0, \dots, M)$ is the least square solution of 
\begin{equation}
    \begin{bmatrix}
        (\Delta T_{k-N})^0 &\! \cdots \! & (\Delta T_{k-N})^M \\
        \vdots &\! \ddots \! & \vdots \\
        (\Delta T_{k+N})^0 & \! \cdots \! & (\Delta T_{k+N})^M \\
    \end{bmatrix}
    \!
    \begin{bmatrix}
        c_{k,0} \\ \vdots \\ c_{k,M}
    \end{bmatrix}
    =
    \begin{bmatrix}
        d[k-N] \\ 
        \vdots \\ 
        d[k+N] 
    \end{bmatrix}. \label{eq:sgfilter}
\end{equation}
Here we set $\Delta T_{k-j} := T[k-j] - T[k]$ and $(\Delta T_{k})^0 = 1$.
We can easily calculate the least square solution by $\hat{\bm{c}} = (A^\top A)^{-1} A^\top \bm{d}$ where $A$ is the $(2N+1) \times M$ coefficient matrix of $(c_{k,0}, \dots, c_{k,M})^\top$ defined in \eqref{eq:sgfilter} and $\bm{d} = (d[k-N], \dots, d[k+N] )^\top$. 
Note that $A^\top A$ is invertible if the sampling timestamp $T[k+j]$ is different for all $j = -N, \dots, N$, because $A$ contains $M\times M$ Vandermonde's matrix as a submatrix and its determinant becomes non-zero.
In the ordinal Savitzky-Golay filter, it is usually assumed that $\Delta T_{k+j} = j$ for all $j = -N, \dots, N$
to reduce a computational cost. In our implementation, 
we decided to recalculate timesteps
to handle with sudden change of sampling rate due to communication issues.
To obtain the derivative of $\omega$, we just differentiate the interpolating polynomial.
\begin{equation}
    \frac{dp}{dt} [k] = \sum_{i=1}^{M} i \cdot c_{k,i} (t - T[k])^{i-1},
\end{equation}
where $c_{k,i}$ is the coefficient of the interpolating polynomial.

If $\bm{d}$ is multidimensional, it can be computed at once by replacing the coefficient vector $\bm{c}$ with a matrix $C$. For example, if we set $\bm{d} = (\bm{\omega}^\top, \bm{a}^\top)^\top$, 
$C$ can be computed as
\begin{equation}
    C = (A^\top A)^{-1} A^\top 
    \begin{bmatrix}
        \bm{\omega}[k-N]^\top & \bm{a}[k-N]^\top \\
        \vdots \\
        \bm{\omega}[k+N]^\top & \bm{a}[k+N]^\top
    \end{bmatrix}.
\end{equation}
Each of column of $C$ corresponds to the each of data column.

Furthermore, thanks to polynomial interpolation, it is possible to match the timestamps with other sensor modules.
Synchronizing the time among sensors is very important in our method, which focuses on the acceleration and angular velocity values at the moment.
We assume here that the each sensor's wall clock is well calibrated, e.g., by using an NTP server, and that only the timestamps of the data are not the same.
The time alignment is illustrated in \figref{fig:interpolation}.



\subsection{Bingham Filter for Relative Orientation Estimation} \label{sec:binghamfilter}

\subsubsection{Principle of Relative Orientation Estimation}
Now we assume that the relative rotation $\vect{R}{A}{P}$ is also unknown.
Assuming that the link is rigid, we can use the classical rigid body theorem that the angular velocity on the body is all the same.
If $i$-th IMU and $j$-th IMU are on the same rigid body, then the following equation holds:
\begin{equation}
    \vect{\gyro}{i}{i} = \vect{R}{i}{j} \vect{\gyro}{j}{j}.
\end{equation}
Now we can estimate $\vect{R}{i}{j}$ from this equation.

%
%
%

\subsubsection{Quaternion and Bingham Distribution}

We try to estimate the probability distribution of $\vect{R}{i}{j}$.
To realize this, we made use of the idea of the Bingham distribution-based linear filter proposed by \cite{binghamfilter}.
The Bingham distribution is the probability distribution over the unit (hyper-)sphere.

The Bingham distribution over 3-sphere is defined as 
\begin{equation}
    \mathfrak{B}(A)(\boldsymbol{q})
    \propto \exp \left(\boldsymbol{q}^{\top} A \boldsymbol{q}\right),
\end{equation}
where $\qt \in \R^4$ is a unit quaternion and $A \in \R^{4\times 4}$ is a parameter of distribution.
Now we define the notation here. 
\begin{enumerate}
    \item The set of all unit quaternion is denoted as $\Sp^3 \subset \R^4$.
    \item We use ``wxyz'' notation.
    \item While $q$ is a quaternion, $\qt$ is the vector in $\R^4$ 
    in the form of $(w,x,y,z)^\top$, where $w$ is a scalar part of $q$.
    \item The \textit{conjugate} of quaternion $q$ is written as $q^*$.
    \item The quaternion product is denoted as $\odot$.
\end{enumerate}



Letting $\bm{v} = (x,y,z)^\top \in \R^3$, we define a tuple notation of quaternion as
\begin{equation}
    q = 
    \begin{pmatrix}
        w \\ \bm{v}
    \end{pmatrix}
    =
    \begin{pmatrix}
        w \\ (x,y,z)^\top
    \end{pmatrix}
    =
    (w, \bm{v})^\top.
\end{equation}
Now we define the multiplication of quaternions $\odot$ as follow:
\begin{equation}
    \begin{pmatrix}
        w\\ 
        \bm{v}
    \end{pmatrix} \odot 
    \begin{pmatrix}
        w' \\ \bm{v}'
    \end{pmatrix} :=
    \begin{pmatrix}
        ww' - \bm{v}^\top \bm{v}' \\
        w \bm{v}' + w' \bm{v} + \bm{v} \times \bm{v}'
    \end{pmatrix}. \label{eq:quat_mult}
\end{equation}

It is well known that quaternions can be express the 3D spatial rotation
\cite{Gallier2011}. 
If we set the conversion function from unit quaternion to rotation matrix $R: \Sp^3 \to \SO(3)$, we get
\begin{align}
    \qt \odot (0, \bm{x})^\top \odot \qt^* =
    (0, R(\qt) \bm{x})^\top \label{eq:quat_and_rot_corresp}
\end{align}

\subsubsection{Bingham Linear Filter}

From \eqref{eq:quat_and_rot_corresp}, we can yield
\begin{equation}
    \qt \odot (0, \bm{b})^\top = (0, \bm{a})^\top \odot \qt,
\end{equation}
where $\bm{a} \in \R^3$ is a point after the rotation and $\bm{b} \in \R^3$ a point before the rotation. Using \eqref{eq:quat_mult}, we get the following.
\begin{equation}
    \underbrace{\begin{bmatrix}
        0 & (\bm{a} - \bm{b})^\top \\
        -(\bm{a} - \bm{b}) & [(\bm{a} + \bm{b}) \times]
    \end{bmatrix}}_{H(\bm{a}, \bm{b})} \qt = \bm{0}. \label{eq:rotation_q_eq}
\end{equation}
Note that all $\qt$ that satisfy \eqref{eq:rotation_q_eq} also satisfy $\bm{a} = R(\qt) \bm{b}$ and vice versa.

The above discussion was for the case where both $\bm{a}$ and $\bm{b}$ have no noise.
In practice, there is observational noise, so we need to consider it. 
Based on \cite{binghamfilter}, separating the noise term from the true value term, then we have
\begin{equation}
    H(\bm{a} + \bm{\varepsilon}_a, \bm{b} + \bm{\varepsilon}_b) \qt
    = H(\bm{a}, \bm{b}) \qt + \bm{\varepsilon}_H.
\end{equation}
%
The mean of $\bm{\varepsilon}_H$ is zero. Under the assumption that $\bm{\varepsilon}_a$ and $\bm{\varepsilon}_b$ are uncorrelated, we can compute its covariance matrix $\Sigma^H$ by
\begin{equation}
    \Sigma^H = {N}\left(
        \begin{bmatrix}
            \bm{\varepsilon}_a & \\
            & \bm{\varepsilon}_b
        \end{bmatrix}
    \otimes {\Sigma}^{\bm{q}}\right) {N}^{\top},
\end{equation}
where $N$ is a $4 \times 24$ matrix defined by
\begin{equation}
    \left[
    \begin{array}{@{\hspace{0mm}}c@{\hspace{1mm}}c@{\hspace{1mm}}c@{\hspace{1mm}}c@{\hspace{1mm}}c@{\hspace{1mm}}c@{\hspace{0mm}}}
        H(\bm{e}_1, \bm{0}) & H(\bm{e}_2, \bm{0}) & H(\bm{e}_3, \bm{0}) & H(\bm{0}, \bm{e}_1) & H(\bm{0}, \bm{e}_2) & H(\bm{0}, \bm{e}_3)
    \end{array}
    \right].
\end{equation}
Here $\bm{e}_i$ is a vector whose $i$-th element is $1$ and others $0$.
The matrix ${\Sigma}^{\bm{q}}$ is the covariance matrix of rotation. We utilize for ${\Sigma}^{\bm{q}}$ the covariance of the uniform distribution over the entire quaternion space $\Sp^3$. It can be written as ${\Sigma}^{\bm{q}} = 0.25 I_4$, where $I_4$ is the identity matrix of size $4$.

Now we formulate the sequential update formula of the Bingham distribution.
First, calculate
\begin{equation}
    \Delta A_{i,j} = -\frac{1}{2} H(\gyro_i, \gyro_j)^\top (\Sigma^{H})^{-1} H(\gyro_i, \gyro_j)
\end{equation}
to obtain an updater. 
If one assumes $\Sigma^{\bm{\varepsilon}}$ is fixed during the task, then $(\Sigma^{H})^{-1}$ can be precomputed and one can save the computational cost.

Letting $\gamma_\text{rot} \in (0, 1]$ be a forgetting factor, we can update the parameter matrix with $\Delta A_{i,j}$ as
\begin{equation}
    A_{i,j}[n+1] = \gamma_\text{rot} \cdot A_{i,j}[n] + \Delta A_{i,j}. \label{eq:binghammatrix}
\end{equation}
Here $A_{i,j}$ is the parameter matrix of the Bingham distribution of $\vect{R}{i}{j}$ and initialized with zero matrix: $A_{i,j}[0] = O_{4\times 4}$.

We can extract the mode quaternion by eigendecomposition. The eigenvector of $A[n]$ corresponding to the maximum eigenvalue is the mode quaternion of the Bingham distribution defined by $\bingham(A[n])$. We use the mode rotation as $\vect{R}{i}{j}$.


The approach to pose estimation using the Bingham distribution does not require a linearization operation and the update rule is simple. 
In addition, it has the advantage of handling global information over the entire rotation space.


    
    

\subsection{Brief Calibration of IMU sensors} \label{sec:calibration}



Before estimating the relative posture of the joints, a simple calibration of the IMU was performed. This is an important operation because the centrifugal force value is critical in this method.

This process is performed as follows.
First, the robot is kept stationary and approximately 1000 samples are collected.
During the stationary state, the bias of the gyro sensor and the covariance between the gyro sensor, its derivative, and the acceleration are measured.
The bias and covariance of the sensors were assumed to be constant during operation, independent of angular velocity or acceleration, and these values were used as constants.

Next, while we slowly rotated the robot, approximately 10000 samples were observed and used to calibrate the accelerometers.
Calibration of the accelerometers is performed by the ellipsoid fitting method, as presented in \cite{imucalibration}. This method is based on the assumption that the magnitude of the gravitational acceleration is constant independent of the orientation of the IMU.
The reason for turning the robot slowly is to ensure that, as far as possible, acceleration other than gravitational acceleration is not observed.
If the robot is fixed to the environment and difficult to rotate, pre-calibrated IMU sensors can be used.

After collecting the data, the calibration results are obtained by the least-squares method.
If we observe an acceleration dataset $\{(X_i,Y_i,Z_i)\}_{i=1}^N$, these points should be on the ellipsoid represented by
\begin{equation}
\begin{aligned}
    a X^2 + b XY + c XZ &+ d Y^2 + e YZ + f Z^2 \\
    & + p X + q Y + r Z + s = 0
\end{aligned}
\end{equation}
where $a,\dots , f, p \dots, s$ are real numbers. With a dataset of observation
, we can construct linear square least problem that find $a,\dots , f, p \dots, s$ minimizing
\begin{equation}
    \underbrace{\left[
    \begin{array}{@{\hspace{-0ex}}c@{\hspace{1ex}}c@{\hspace{1ex}}c@{\hspace{1ex}}c@{\hspace{1ex}}c@{\hspace{1ex}}c@{\hspace{1ex}}c@{\hspace{1ex}}c@{\hspace{1ex}}c@{\hspace{1ex}}c@{\hspace{-0ex}}}
        X_1^2 & X_1 Y_1 & X_1 Z_1 & Y_1^2 & Y_1 Z_1 & Z_1^2 & X & Y & Z & 1 \\
        \multicolumn{10}{c}{\vdots} \\
        X_N^2 & X_N Y_N & X_N Z_N & Y_N^2 & Y_N Z_N & Z_N^2 & X & Y & Z & 1
    \end{array}
    \right]}_{D_{1:N}} \hspace{-1ex}
    \underbrace{\begin{bmatrix}
        a \\ 
        \vdots \\ s
    \end{bmatrix}}_{\bm{v}}.
\end{equation}

Now we have the optimal parameter of the quadratic cone.
However, it does not necessarily ellipsoid.
In practice, we restricted our estimation results to ellipsoid surfaces that are not hyperbolic or parabolic, based on \cite{ellipsoid_1290055}.

\section{EXPERIMENTS}

\subsection{Validation Experiment}

\begin{figure}[t]
    \centering 
    \includegraphics[width=0.9\linewidth]{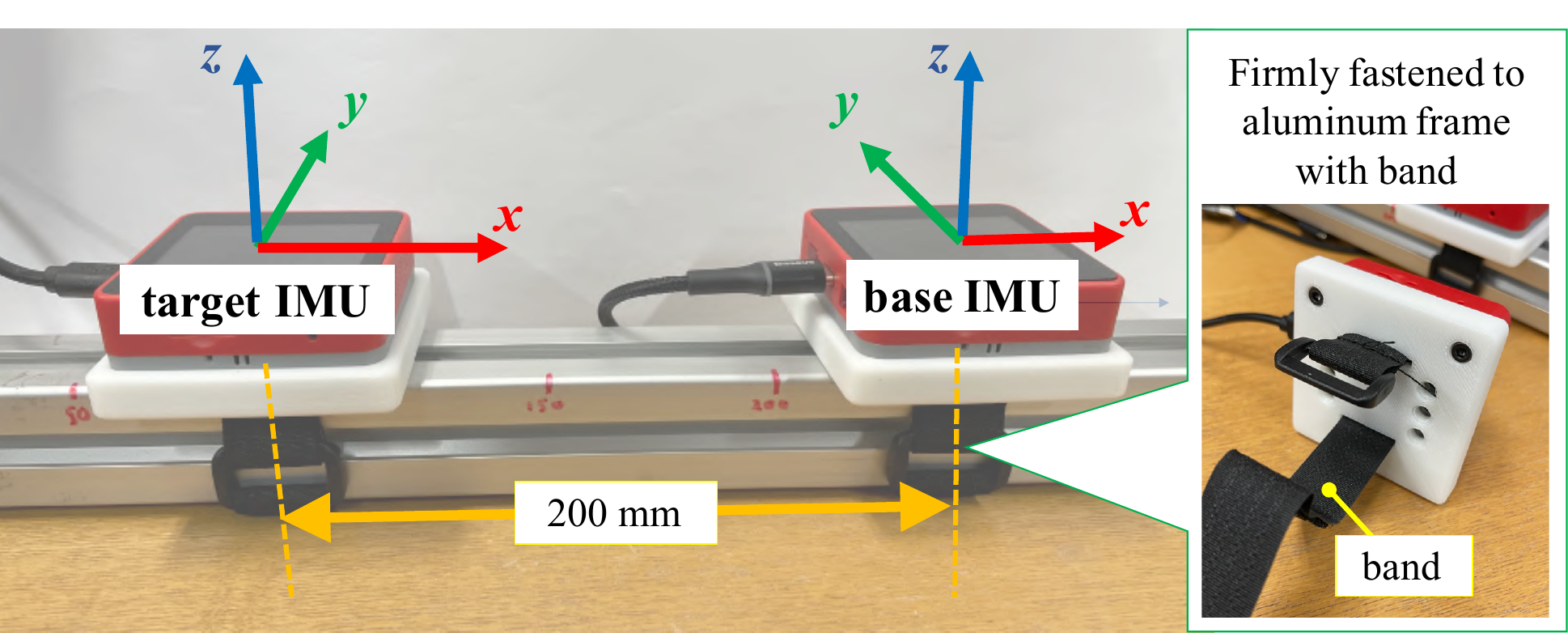}
    \caption{Setup of the validation experiment. Two M5Stack Fire were used as the IMU modules. The IMU modules were fixed to aluminum rods with 200 mm spacing on the rods.}
    \vspace{-0.25cm}
    \label{fig:200mm_test_photo}
\end{figure}

\begin{figure}[t]
    \centering
    \includegraphics[width=\linewidth]{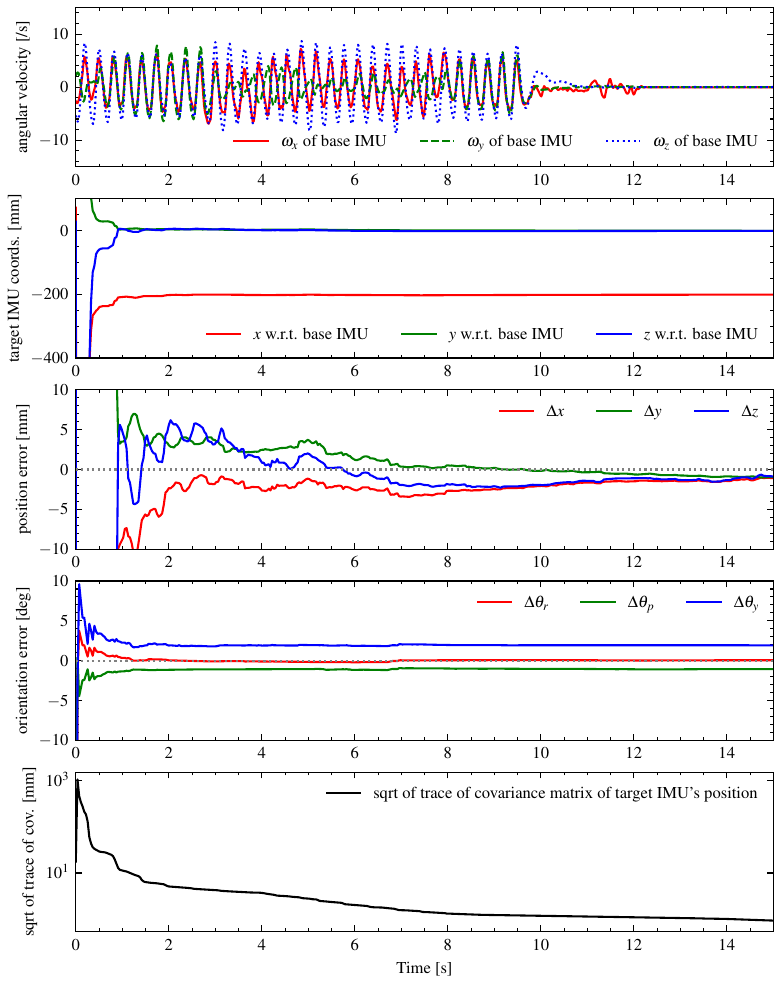}
    \caption{Results of validation experiments. From the top row, the time variation of the angular velocity during the experiment, the estimated relative position, the position error, the orientation error, and the square root of the trace of covariance of the estimation results are depicted.}
    \vspace{-0.25cm}
    \label{fig:200mm_test}
\end{figure}

As a validation experiment, we used a pair of IMU sensor modules strapped to an aluminum rod of known length.
\figref{fig:200mm_test_photo} shows the setup used in the evaluation experiment.
The modules were aligned to the same orientation.
To verify that the relative pose was estimated correctly, we shook the rods by hand. In this section, we set $\gamma_\text{pos} = \gamma_\text{rot} = 1$.

\figref{fig:200mm_test} shows the estimation results.
The position estimation error is at most about 3 mm and the orientation error is at most about 3${^\circ}$, which are good results. 
The square root of the trace of the covariance matrix of the estimation results is about 1 mm at the end.
By analogy with the normal distribution, three times this value should be approximately the 99.7\% percentile, but this value is seemed to be an underestimate considering the actual error. 
One possible reason for the underestimation is that the assumption of least-squares regression that the error distribution is Gaussian is not strictly met.
Although the assumption is not being followed, we formally utilize the trace of the covariance matrix as a termination condition for the estimation.



\begin{figure}[t]
    \centering
    \includegraphics[width=0.9\linewidth]{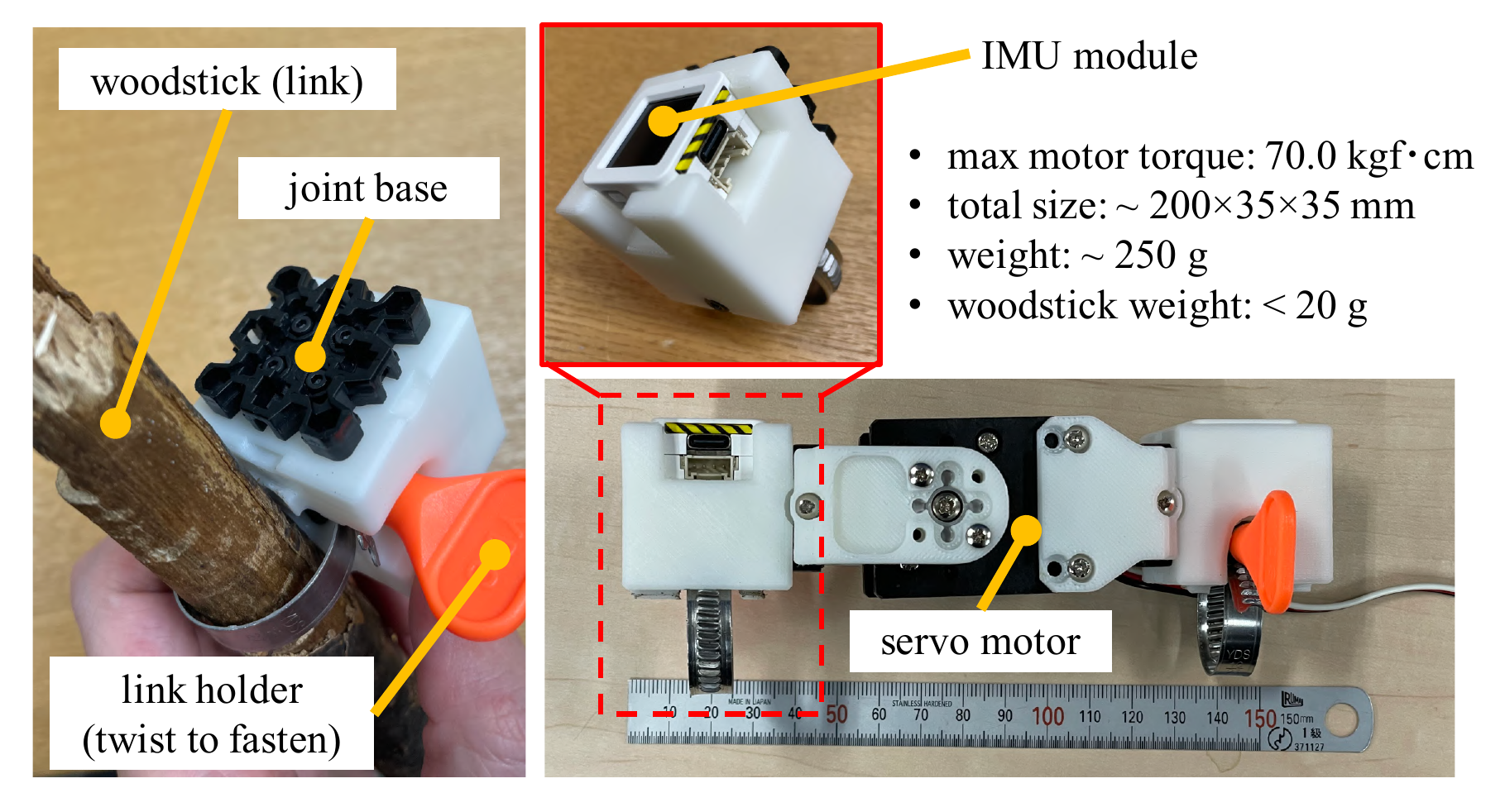}
    \vspace{-0.25cm}
    \caption{Our joint module used in the application experiments.}
    \vspace{-0.25cm}
    \label{fig:joint_module}
\end{figure}


\subsection{Design of the Joint Module Used in the Application Experiment}

We developed a joint intermediate module for on-site built robots that have:
\begin{enumerate}
    \item a structure to fix the link and the module,
    \item an actuator (e.g. servo motors),
    \item two IMU sensors. \label{item:imu}
\end{enumerate}
Requirement \ref{item:imu} is needed to simplify the estimation problem.
One of the two sensors is fixed to the servo motor and the other to the servo horn (child side of the drive shaft).
This arrangement allows the two IMU sensors to be mounted on the same rigid link and the proposed method can be used.

The appearance and specifications of the joint intermediate module are shown in \figref{fig:joint_module}.
The module consists of joint parts, a pair of IMU modules, a servo motor, and 3D printed structures.
The motor on the root side has enough power to drive the entire arm.
A small, lightweight motor is used on the tip side to reduce the motor torque on the root side.

Since the position of the IMU with respect to the joint module is known at the design stage, the relative position of the IMUs relative to the joint can be given in advance. Using this information, the relative pose between the joints can be calculated from the relative pose between the IMUs.

\subsection{Application Experiment}

\begin{figure*}[t]
    \centering
    \includegraphics[width=0.925\linewidth]{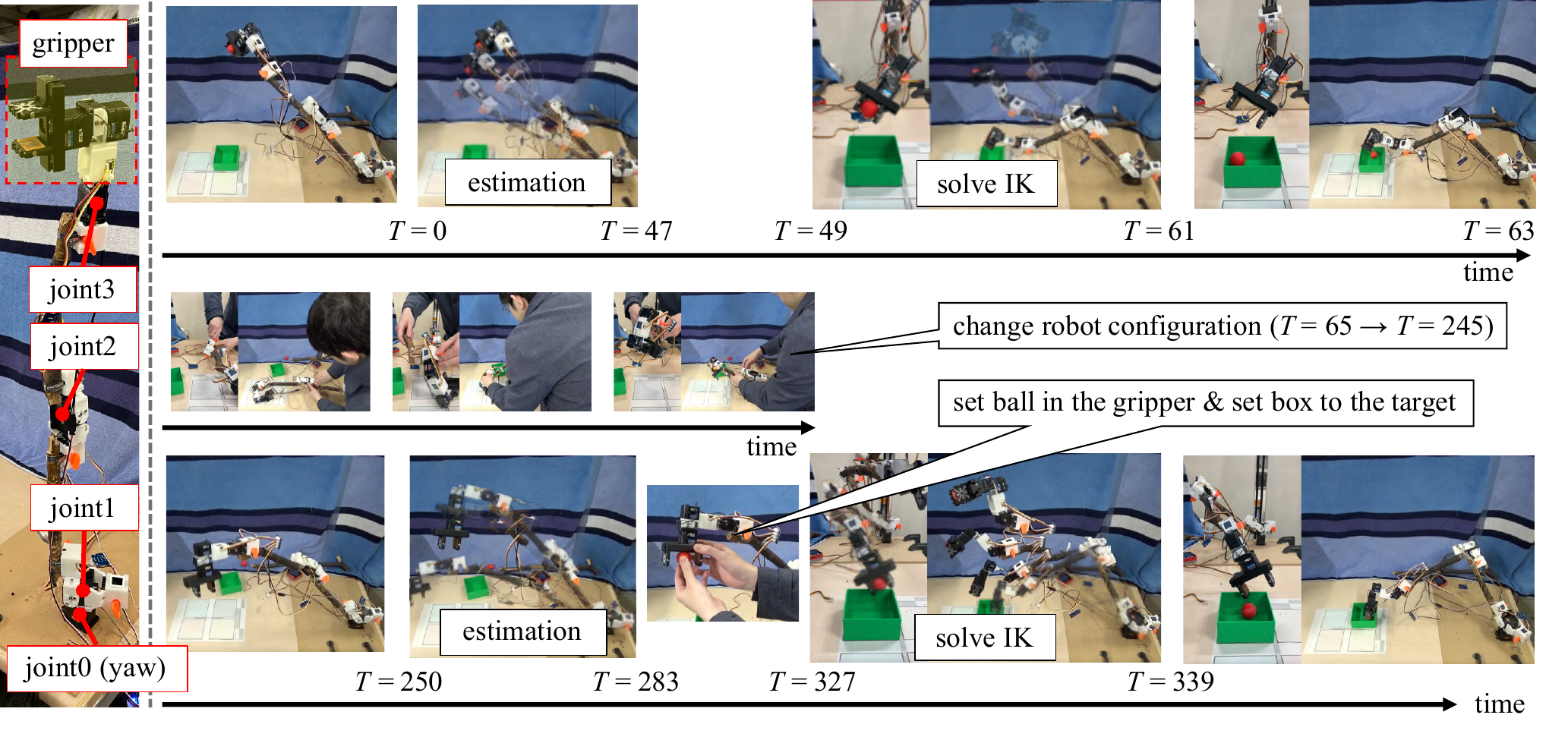}    \vspace{-0.33cm}
    \caption{Snapshot of an application experiment. The reaching motion is considered successful if the red ball can be dropped into the green box. Although the configuration of the robot was changed during $T=65 \to T=245$, reaching was successful both before and after the change. The joint configuration of the robot is illustrated on the left side of the figure. ``joint1'' to ``joint3'' can be rotated in the pitch direction.}
    \vspace{-0.33cm}
    \label{fig:applexp_photo}
\end{figure*}

\begin{figure}[t]
    \centering
    \includegraphics[width=\linewidth]{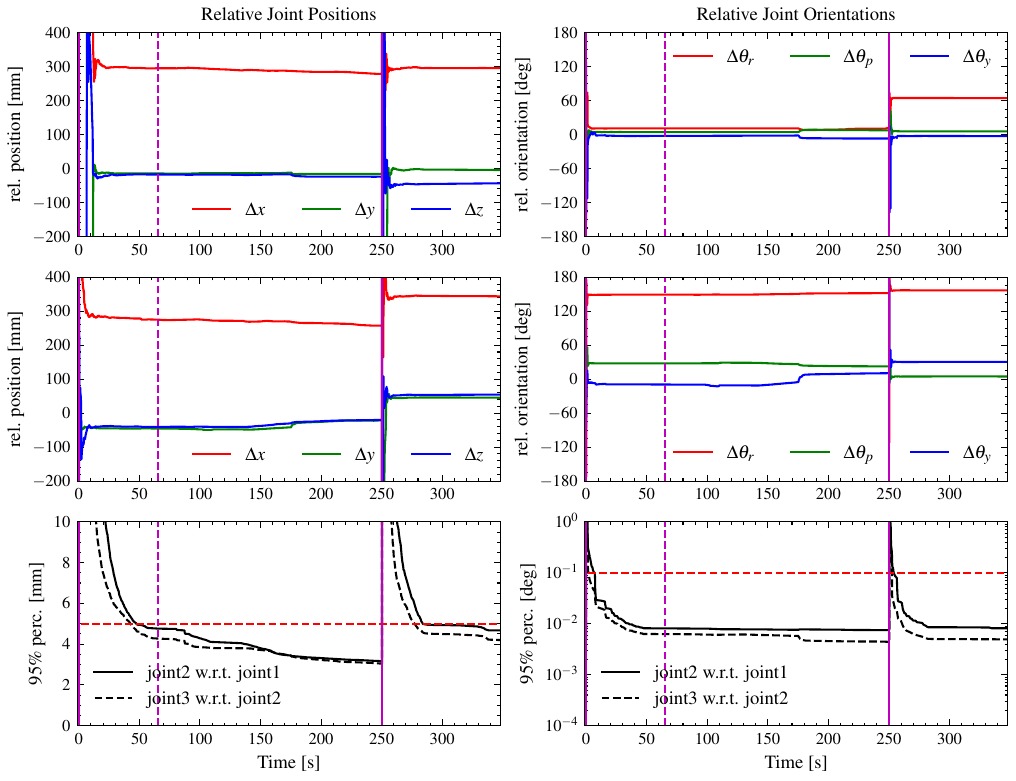}
    \caption{Estimation results during an application experiment.
    From the top row, time variation of the relative pose of joint1 to joint2, the relative pose of joint2 to joint3, and the 95\% percentile (twice the square root of the trace of the covariance matrices) the pose variance.
    The vertical dashed line in magenta is the time when the robot started to change its configuration. The solid magenta vertical line is the time when the estimated motion began.}
    \vspace{-0.33cm}
    \label{fig:applexp_data}
\end{figure}

We show that a manipulator made of wooden sticks can reach a gripper to a target position with our estimation method.
The wooden stick used was a cherry tree (Someiyoshino), which was picked up by the author on our university campus.
Reaching was assumed successful by dropping a red ball ($\phi$20 mm) into a placed green box (90 mm $\times$ 90 mm $\times$ 40 mm).
The position of the green box was given in advance relative to the robot's reference coordinates. No visual feedback was used, and only the estimated link length was used for the inverse kinematics of the gripper.

The experiment is shown in \figref{fig:applexp_photo} and the estimation result \figref{fig:applexp_data}.
As soon as the estimation motion started ($T=0$), a signal was sent to start estimation and the previous estimation result was reset.
The sampling rate of IMUs was 85 Hz.

A swinging signal was sent to a randomly selected joint at a fixed angle amplitude.
The estimation process stopped automatically ($T=47$) when the covariance fell below a threshold, which was set at 5 mm at ``the 95th percentile'', and the orientation at 0.1 deg at ``the 95th percentile''.

When the estimation motion was completed, the robot reached the gripper to the target box based on the estimated kinematic model. Note that although the actual robot deflected due to gravity, this effect was assumed to be negligible in this experiment because only the horizontal plane was focused on in this experiment. To prevent collision with the box due to deflection, the offset was set 50 mm above the box.
When the gripper reached the target position, it automatically opened and dropped the ball ($T=63$). As shown in \figref{fig:applexp_photo}, the robot successfully dropped the red ball into the green box.

Next, the servo motors of the robot were turned off, the robot was disassembled, and the configuration was changed.
We replaced the wooden sticks with different ones and changed the mounting direction drastically of the joint modules.
At this time, the robot was manually moved like a puppet to make sure the gripper could reach the green box.

The motors were turned on again and estimation was started at time $T=250$. The estimation was completed ($T=283$) and the gripper moved over the target box ($T=339$) and successfully dropped the ball into the box.
Comparing the final posture of the robot when the ball was dropped, it can be seen that the angles of each joint were very different. This indicates that the robot solved the inverse kinematics again based on the estimated 
joints' relative poses.

\section{DISCUSSIONS}

\subsection{Effect of Sampling Frequency of IMUs}
\label{sec:freqanalysis}

Our method uses a Savitzky-Golay filter to smooth and interpolate the sensor data and calculate the derivative values.
There is concern that this frequency characteristics may affect the estimated values.
The characteristics is determined by the ratio of the shake frequency to the sampling frequency.

To discuss the effect of sampling frequency on estimation, 
simulations were performed assuming 
an IMU sensor placed at relative position $(L,0,0)$ and rotated with angular velocity $(0,0,\sin(2\pi f t))$, where $f$ is a frequency of shaking and $t$ is the time. No Gaussian noise is added here.
%
Throughout the simulation, it was observed that the ratio of the estimated relative position $\hat{L}$ to the true value $L$ and the ratio to swing frequency $f$ to sampling frequency $f_\text{sample}$ were constant for the magnitudes of $L$ and $f$.
%
%
Based on this observation, we considered the relationship between the non-dimensionalized values of $\hat{L} / L$ and $f / f_\text{sample}$.
In the simulation, the sampling frequency was 100 Hz.

\begin{figure}[t]
    \centering
    \includegraphics[width=0.95\linewidth]{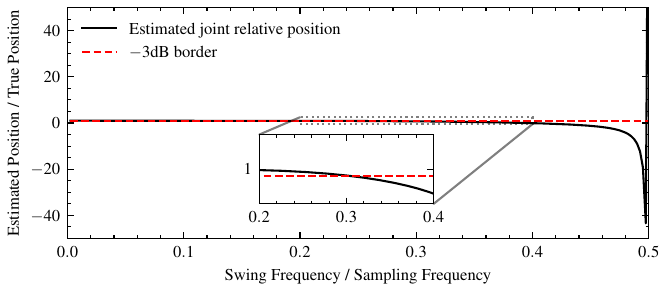}
    \caption{Frequency characteristics of our filter. In our experiments, we set the number of points in the window $N = 3$ and polynomial degree $M=5$.}
    \label{fig:filter_freq}
\end{figure}

\figref{fig:filter_freq} shows the simulation results.
The red dashed line labeled "$-3$dB border" indicates a line whose value is $10^{-3/20} \approx 0.71$. This value follows the fact that the filtering cutoff gain is usually defined at $-3$dB.
The results show that the ratio $\hat{L}/L$ worsens as the ratio $f / f_\text{sample}$ increases. For $N=3$ and $M=5$, the ratio $\hat{L}/L$ falls below the line around $f / f_\text{sample} \approx 0.3$.
This suggests that too fast vibration can cause estimation accuracy to deteriorate.

It is important to note that the way in which this change occurs and the value of the cutoff frequency will differ greatly depending on how $M$ and $N$ are determined. However, in all cases, the higher the frequency, the lower the performance of filtering.
%
%
For a general discussion of the frequency response of the Savitzky-Golay filter, see \cite{sgfilter_freq}.

\subsection{Effect of Estimation Motion}

Let us consider how the estimation motion affect the estimation results.
Noticing that $K_{\overline{\Omega}}(\Sigma^{\vect{\gyro}{A}{A}}, \Sigma^{\vect{\gyro}{P}{P}})$ in \eqref{eq:Omega_zero} is constant if we assume the covariances of sensor noise are unchange during operation,
\begin{equation}
    \overline{\Omega}_0^{(k)} \approx \overline{\Omega}(\vect{\gyro}{A}{A}^{(k)}, \vect{\gyro}{P}{P}^{(k)})
\end{equation}
for a large angular velocity. According to \eqref{eq:P_sequential}, we can see $P[n]^{-1}$ tends to have large entries as $\overline{\Omega}_0^{(k)}$ increases.
This suggests that the more intense the shaking, the smaller the estimation variance due to the error term is expected to be.


\subsection{Limitation and Future Works}

In our method, we use a pair of IMU sensors installed on the same rigid link to estimate the relative pose between the IMUs.
If one tries to estimate the pose of the end effector directly, the entire robot may vibrate, and the rigid body assumption may not be satisfied.
In the verification experiments, we estimated the length of each link one by one. 
However, even in cases where the rigid body assumption fails locally,
it would be possible to further improve performance by making full use of sensor information, such as by applying a graph-based SLAM method \cite{graphbased-slam}.

In this paper, we focused on the estimation of the kinematic model. 
In order to generate the behavior of more complex robots,
a dynamic model should be required, whose estimation is our future work.


\section{CONCLUSION} 

In this paper, we proposed a method for estimating the kinematic model of a robot composed of arbitrary rigid links using an IMU sensor module for proof of concept of on-site building robotics.
We created a manipulator made up of wooden sticks using the designed joint intermediate module.
Even after making significant changes to the robot's configuration, we confirmed that the robot was able to reach the object successfully, which suggests that the robot was able to correctly estimate its kinematic model by our method.






\bibliographystyle{IEEEtran}
\bibliography{IEEEabrv,root}
\end{document}